\documentclass[11pt]{article}

\usepackage[final]{acl}

\usepackage{times}
\usepackage{latexsym}

\usepackage[T1]{fontenc}

\usepackage[utf8]{inputenc}

\usepackage{microtype}

\usepackage{inconsolata}

\usepackage{graphicx}

\usepackage{kotex}
\usepackage{amsmath}
\usepackage{multirow}
\usepackage{arydshln}
\usepackage[normalem]{ulem}
\useunder{\uline}{\ul}{}
\usepackage{tcolorbox}
\usepackage{caption}
\usepackage{subcaption}
\usepackage{booktabs}
\usepackage{colortbl}
\usepackage{fixltx2e}
\usepackage{setspace}
\usepackage{tabularx}
\usepackage{alltt}
\usepackage{listings}
\usepackage{amssymb}
\usepackage{pifont}   
\usepackage{xcolor}   
\usepackage[dvipsnames]{xcolor}
\usepackage[table]{xcolor}
\usepackage{algorithm}
\usepackage{algpseudocode}
\usepackage{subcaption}
\usepackage{float}
\usepackage{adjustbox}
\usepackage{amsthm}
\usepackage{makecell}

\newtcbox{\boxedtag}{on line,
  colframe=Blue,    
  colback=Orange!58, 
  boxrule=0.3pt,     
  arc=0pt,           
  boxsep=0.1pt,        
  left=1pt,
  right=1pt,
  top=0.5pt,
  bottom=0.5pt
}

\title{EXPO-SQL: Execution-based Clause-level Policy Optimization for Text-to-SQL}

\author{
 \textbf{Jaehoon Lee},
 \textbf{CheolWon Na},
 \textbf{Suyoung Bae},\\
 \textbf{Jin-Seop Lee},
 \textbf{Jihyung Lee},
 \textbf{YunSeok Choi}\thanks{Corresponding authors},
 \textbf{Jee-Hyong Lee}$^*$
\\
 College of Computing and Informatics \\ Sungkyunkwan University, South Korea
\\
  \{hoon1223, ncw0034, sybae01, wlstjq0602, jjklle, ys.choi, john\}@skku.edu
}

\begin{document}
\maketitle
\begin{abstract}




Text-to-SQL enables users to query databases using natural language by generating executable SQL queries.
Recent methods have increasingly adopted Large Language Models based reinforcement learning (RL) to leverage execution feedback for training.
However, existing RL methods assign uniform query-level rewards to all clauses in a SQL query, treating correct and incorrect clauses equally. 
This coarse-grained reward design leads to insufficient learning signals for correct SQL generation.
To address this issue, we propose \textbf{EXPO-SQL} (\textbf{EX}ecution-based clause-level \textbf{P}olicy \textbf{O}ptimization for Text-to-\textbf{SQL}) which provides fine-grained supervision through clause-level rewards. 
To assign clause-level rewards, our method identifies erroneous clauses by analyzing execution results, including error messages and clause-wise incremental execution.
Experiments on widely-used Text-to-SQL benchmarks demonstrate that EXPO-SQL significantly outperforms existing supervised fine-tuning, prompting, and RL-based methods through fine-grained clause-level learning.
Our code is available at \url{https://github.com/jhn25/EXPO-SQL}.

\end{abstract}

\section{Introduction}

Text-to-SQL aims to generate executable SQL queries from given natural language question and database schema.
It is a core technology that facilitates data retrieval by enabling non-expert users to query databases directly~\citep{nltodbsurvey,nltodbsurvey2}.
%
Recent works on Text-to-SQL has increasingly adopted large language models (LLMs), leveraging their strong reasoning ability~\citep{DBLP:journals/corr/abs-2408-05109,recentsurvey,death,hong2024next}.

Early approaches adopted supervised fine-tuning (SFT) to optimize token-level matching with gold SQL queries~\citep{xiyansql,dtssql}.
Prompt-based methods leverage in-context learning with iterative refinement at inference time~\citep{xiyansql,dtssql}. 
However, both approaches still produce erroneous SQL queries due to the lack of execution-aware supervision for Text-to-SQL reasoning.
\begin{figure}[t]
  \includegraphics[width=\columnwidth]{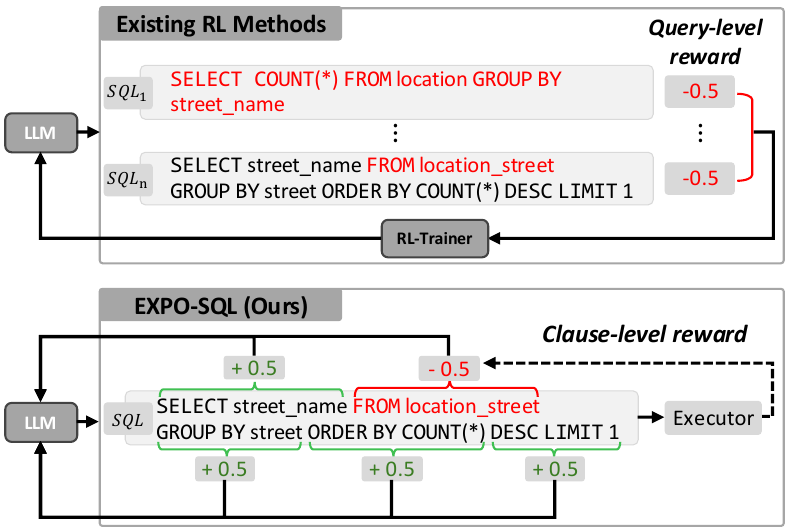}
  \caption{Comparison of reward assignment strategies. Existing RL methods assign query-level rewards uniformly across all clauses, while EXPO-SQL assigns clause-level rewards, only penalizing erroneous ones.}
  \vspace{-0.4cm}
  \label{fig:intro}
\end{figure}

To address this issue, several studies have explored reinforcement learning (RL) approaches for Text-to-SQL generation~\citep{sqlrlsurvey,sqlrlmapo,sqlsurveyrl2,sqldpo}.
These methods optimize model policies using execution results as reward signals~\citep{sqlr1, arctic, excot, reasoningsql, graphrewardsql, rewardsql}.
However, these RL-based approaches have a limitation that they rely on query-level learning signals. 

Usually, an SQL query is composed of several clauses, such as \texttt{SELECT, FROM, WHERE}, etc. In practice, execution failures are often caused by errors in a few clauses rather than the entire clauses in the SQL query. However, existing methods assign an identical reward to all clauses in a generated SQL query, as illustrated in Figure~\ref{fig:intro} where an identical reward is given to all clauses even when the \texttt{FROM} clause is only incorrect.
This query-level rewards treat correct and incorrect clauses equally,
leading to coarse credit assignment~\citep{ca} and penalizing even correctly generated clauses.
As a result, the model receives only coarse learning signals, which are insufficient for correct SQL generation.

To address this coarse credit assignment problem, we propose \textbf{EXPO-SQL} (\textbf{EX}ecu\-tion-based clause-level \textbf{P}olicy \textbf{O}ptimization for Text-to-\textbf{SQL}), which separately evaluates the correctness of each clause within a SQL query and provides fine-grained supervision with clause-level reward. 
%
However, evaluating each clause remains highly challenging in online reinforcement learning. 
First, lexical analysis of SQL query cannot determine which clauses have error.
Second, due to the one-to-many nature of SQL queries, whereby various queries can produce the same execution result, direct token matching with the gold SQL is not appropriate. 
Finally, execution results only provide binary feedback on whether the entire query is correct.

To overcome these limitations, we analyze execution results to identify erroneous clauses, rather than using them directly as binary rewards. 
We first classify them into three cases and design corresponding strategies: correct results, incorrect results, and execution errors.
In the case of {\it correct results} where the query is executed successfully and produces the correct answer, we assign positive rewards to all clauses. 
For {\it incorrect results}, where the query is executable but produces an incorrect answer, we decompose the generated SQL query following the logical execution order and incrementally execute each clause. Then, we analyze how the result changes before and after adding each clause to identify erroneous clauses. 
Moreover, in the case of {\it execution errors} where the query fails to be executed, we analyze the error message to identify clauses that caused the failure.

For each case, we analyze how each clause affects the result and design clause-level rewards that provide fine-grained learning signals. 
Through the execution-based analysis, EXPO-SQL provides more accurate clause-level learning signals, which existing RL methods with query-level rewards cannot achieve.

We evaluate the superiority of EXPO-SQL on widely-used Text-to-SQL benchmarks including Spider~\citep{spider} and BIRD~\citep{bird}, comparing with a wide range of baselines including SFT, prompting, and recent RL-based methods.
Experimental results demonstrate that EXPO-SQL significantly improves execution accuracy through effective clause-level learning signals, achieving state-of-the-art performance. Specifically, EXPO-SQL outperforms the best RL baseline by 1.2\%p on Spider-Dev and 2.4\%p on BIRD-Dev. The improvements are more pronounced on complex queries, showing 5.6\%p gains.

\section{Related Work}
\subsection{Conventional Text-to-SQL Methods}
Early Text-to-SQL methods fine-tuned models using large-scale text-SQL pair datasets~\citep{SENSE, dtssql, codes, omnisql}, with recent work incorporating chain-of-thought reasoning~\citep{cot} through schema linking and query decomposition~\citep{wang-etal-2025-linkalign, route}.
With the emergence of LLMs, prompting-based methods have been extensively studied, including in-context learning with well-designed demonstrations~\citep{dailsql, dinsql, dcgsql}, SQL-specific chain-of-thought prompting~\citep{c3sql, dividenprompt}, and execution consistency based selection strategies with multiple generated candidates ~\citep{mcssql}.
However, these methods did not leverage execution feedback as learning signals. 

\subsection{Reinforcement Learning in Text-to-SQL}
Recently, reinforcement learning has been applied to Text-to-SQL leveraging execution feedback~\citep{sqlr1, reasoningsql, arctic}. Most methods adopted GRPO~\citep{grpo} for policy optimization ~\citep{arctic,sqlr1,rewardsql,reasoningsql}, while some explored DPO~\citep{dpo} by constructing preference pairs from execution results~\citep{excot}.

\begin{figure*}[t]
    \centering
    \includegraphics[width=\linewidth]{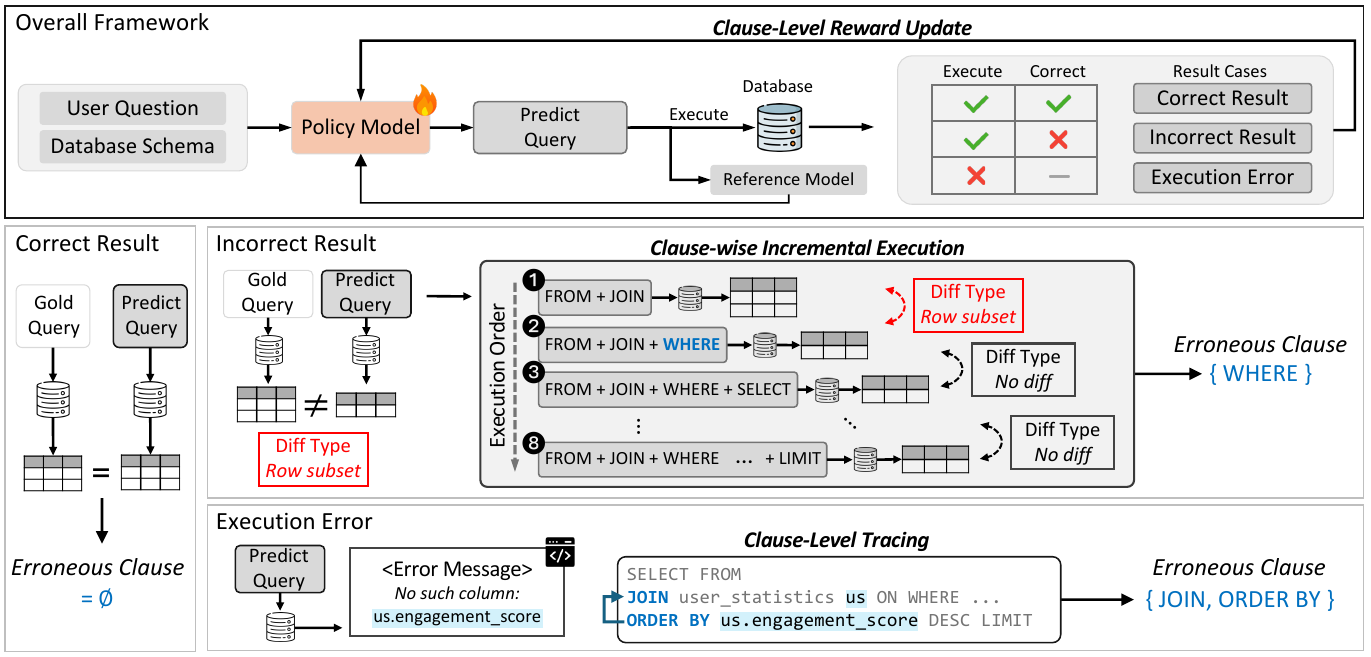}    
    \caption {The overall framework of EXPO-SQL. Predicted queries are classified into three cases based on execution results and correctness. For each case, erroneous clauses are detected to assign clause-level rewards: correct results have no erroneous clauses (Section~\ref{correct}), incorrect results use difference type detection through incremental execution (Section~\ref{result_error}), and execution errors employ error messages and clause-level tracing (Section~\ref{exec_error}).}
    \label{mainfig}
\end{figure*}

Since execution results provide only binary feedback, some methods designed additional rewards for more informative signals. 
ReasoningSQL~\citep{reasoningsql} combined LLM-as-a-judge evaluation with syntactic validity and schema matching rewards. RewardSQL~\citep{rewardsql} introduced a process reward model that evaluates intermediate reasoning steps. GraphRewardSQL~\citep{graphrewardsql} proposed the graph-based reward with gold SQL comparison.

However, those methods assigned rewards at the query-level, treating all clauses equally despite only specific clauses being erroneous. They failed to distinguish correct clauses from erroneous ones, resulting in insufficient learning signals.

\section{EXPO-SQL}
In this section, we introduce EXPO-SQL, a novel clause-level reinforcement learning framework. Figure~\ref{mainfig} illustrates the overall framework of EXPO-SQL.
Instead of using execution feedback as a query-level reward, we utilize execution results to identify clause-level errors and design clause-level rewards.
%
We first formulate Text-to-SQL generation as clause-level policy optimization in Section~\ref{3.1}. We then present execution-based clause-level reward design for different execution results in Section~\ref{3.2}, and describe how these rewards are incorportaed into policy optimization in Section~\ref{3.3}.



\subsection{Problem Formulation} \label{3.1}
Given a natural language question and a database schema, the policy model $\pi_\theta$ generates a SQL query $O$. We define the generated SQL query as a sequence of $n$ clauses: $(c_1, c_2, \ldots, c_n)$, where each clause corresponds to a structural component of the SQL query.
The generated query is executed against the database to obtain an execution result table $R_{pred}$, then the reward signal is extracted by comparing $R_{pred}$ with ground-truth $R_{\text{gold}}$.

Our goal is to provide clause-level learning signals, rather than query-level rewards, by assigning differentiated rewards $r_c$ to individual clauses $c \in O$.
To this end, we erroneous clauses $C_{\text{err}} \subseteq \{c_1, \ldots, c_n\}$ by leveraging detailed execution feedback, including error messages and differences between execution results. Based on the identified erroneous clauses, we design clause-level rewards $r_c$ that guide the policy model toward improved SQL generation.



\subsection{Clause-level Reward Design} \label{3.2}
In order to adopt fine-grained clause-level rewards, we identify erroneous clauses by analyzing the execution result of the generated SQL query. Since the information available for identifying erroneous clauses varies depending on execution results, it is necessary to apply different approaches accordingly. 
Therefore, we categorize execution results into the following three cases:
(1) Correct Result: the query is executed successfully and $R_{pred} = R_{gold}$. (2) Incorrect Result: the query is executable but $R_{pred} \neq R_{gold}$. (3) Execution Error: the query fails to be executed.
For each case, we effectively identify $C_{err}$ by leveraging detailed execution feedback and assign different rewards $r_c$ to each clause $c \in O$ to provide fine-grained learning signals. 

\subsubsection{Correct Result} \label{correct}
When the generated SQL query is successfully executed and produces the correct result, all clauses are considered correct, meaning $C_{err}=\emptyset$. Therefore, we assign the same positive reward to all clauses.
\begin{equation}
r_c = +1.5 \quad \forall c \in O
\end{equation}

\subsubsection{Incorrect Result} \label{result_error}
When the generated SQL query is executable but the execution result is incorrect, there is no direct information about which clause caused the error. 
In this case, to identify $C_{err}$, we consider analyzing the difference between $R_{pred}$ and $R_{gold}$.
However, since both results show the tables with the cumulative effect of all clauses, we cannot directly identify which clauses lead to the incorrect result.
To address this challenge, we leverage each SQL clause performs a distinct operation on the result.
By analyzing these clause-specific effects, we can trace back which clause caused the differences in the final result. 
%
We accomplish this by two steps. First, we define \textit{difference types from execution results} to specify which difference each clause can produce. Second, with those types, we perform \textit{clause-wise incremental execution} to detect the actual effect caused by each clause, thereby identifying $C_{err}$.

\begin{table}[t]
\centering
\resizebox{0.97\columnwidth}{!}{
\begin{tabular}{c|l|l}
\toprule
\textbf{Level}& \textbf{Diff Type} & \textbf{Description} \\
\midrule
\multirow{2}{*}{Column}&\texttt{col\_count} & Number of columns differs \\
&\texttt{col\_name} & Column names differ \\
\midrule
\multirow{7}{*}{Row}&\texttt{row\_order} & Row sequence differs \\
&\texttt{row\_dedup} & Duplicate counts differ \\
&\texttt{row\_subset} & Some rows are missing \\
&\texttt{row\_superset} & Extra rows are present \\
&\texttt{row\_emptied} & Result is empty, gold is not \\
&\texttt{row\_created} & Result has rows, gold is empty \\
&\texttt{row\_disjoint} & No intersection between results \\
\midrule
Value & \texttt{row\_partial} & Cell values differ \\
\bottomrule
\end{tabular}
}
\caption{The 10 difference types (Diff Type) based on changes in columns, rows, values.}
\label{tab:diff_types_main}
\end{table}

\paragraph{Difference Types from Execution Result.}
We first define 10 difference types, denoted as diff\_type from execution result based on the logical operations of SQL clauses.
Since SQL results are returned as tables, any difference between two execution results appear as changes in the columns, rows or values.
Table~\ref{tab:diff_types_main} shows the 10 diff\_type.
More detail explanations are provided in Appendix~\ref{A.1}.




\paragraph{Clause-wise Incremental Execution.}
After defining the diff\_type, we detect $C_{err}$ that caused the diff\_type observed between $R_{pred}$ and $R_{gold}$. Identifying $C_{err}$ is straightforward when each diff\_type is produced by only one clause. However, in most SQL queries, diff\_type can be caused by multiple clauses.
Therefore, we need a comprehensive analysis across the entire generated SQL query to determine how each clause affects the execution result.
\begin{algorithm}[t]
\small
\caption{Clause-wise Incremental Execution}
\label{alg:incremental_main}
\begin{algorithmic}[1]
\Require SQL $O = (c_1, ..., c_n)$, $R_{gold}$
\Ensure Erroneous clauses $C_{err}$
\State $C_{err} \leftarrow \emptyset$
\State Replace \texttt{SELECT} clause with \texttt{SELECT *}
\State $(c'_1, ..., c'_n) \leftarrow$ Reorder$(O)$ in logical execution order
\State $R_1 \leftarrow$ Execute($c'_1$) \Comment{\texttt{FROM/JOIN}}
\State $d_1 \leftarrow$ diff($R_{gold}$, $R_1$)
\For{$i = 2$ to $n$}
    \State $R_i \leftarrow$ Execute($c'_1, ..., c'_i$)
    \State $d_i \leftarrow$ diff($R_{i-1}$, $R_i$)
\EndFor
\State $D^* \leftarrow$ diff($R_{gold}$, $R_{pred}$)
\For{each clause $c'_i$}
    \If{$d_i \in D^*$}
        \State $C_{err} \leftarrow C_{err} \cup \{c'_i\}$
    \EndIf
\EndFor\\
\Return $C_{err}$
\end{algorithmic}
\end{algorithm}
To enable this, we leverage the deterministic execution order of SQL. 
When executing an SQL query, clauses are operated in a predefined logical order, with each clause accessing tables and transforming the previous clause's outcome~\citep{formalsqlvalidation, benzaken2019coq}.
The logical execution order is as follows: \texttt{FROM/JOIN} $\rightarrow$ \texttt{WHERE} $\rightarrow$ \texttt{GROUP BY} $\rightarrow$ \texttt{HAVING} $\rightarrow$ \texttt{SELECT} $\rightarrow$ \texttt{ORDER BY} $\rightarrow$ \texttt{LIMIT}. 
We first decompose the generated SQL query based on the order, and incrementally add clauses one by one starting from the \texttt{FROM/JOIN} clause in the order. 
By observing how the execution result $R_i$ changes as each clause $c_i$ is added, we can identify which clause leads to the difference. Algorithm~\ref{alg:incremental_main} provides the detailed procedure.


%

By comparing the results between the cumulative executions, we can classify the diff\_type $d_i =\text{diff}(R_{i-1}, R_i)$ for each clause $c_i$.
For the first execution where only the \texttt{FROM/JOIN} clause is executed, there is no previous result to compare. In this case, we classify the diff\_type against $R_{gold}$, and only consider whether the diff\_type is \texttt{column\_count}.

After obtaining all diff\_type through clause-wise incremental execution, we also classify the final difference $D^* =\text{diff}(R_{pred}, R_{gold})$, which represents the cumulative effect of all individual clause changes during execution.
A clause $c_i$ is identified as erroneous clause and added to $C_{err}$ if its corresponding diff\_type $d_i$ appears in $D^*$.
More details are provided in Appendix~\ref{A.2}. 
We assign clause-level rewards as follows:

\begin{equation}
r_c = \begin{cases}
-0.5 & \text{if } c \in \mathcal{C}_{\text{err}} \\
+0.5 & \text{otherwise}
\end{cases}
\end{equation}
Since the query is executable with valid syntax, we use relatively weak rewards to fix only $C_{err}$ while preserving the correct clauses. This allows the model to learn which clauses to fix while retaining ones that are already correct.
A detailed case study is provided in Appendix~\ref{C.2}.

\subsubsection{Execution Error} \label{exec_error}
When the query fails to be executed, we leverage the error message from the executor to provide information about $C_{err}$. The message typically contains the error location, type, and the names of tables or columns that caused the error. Details of the error messages are provided in Appendix~\ref{appendix:sqlite_errors}.

However, the clause indicated by the error message is where the execution failed, not necessarily the root cause. As shown in Figure~\ref{mainfig}, while the error message points to \texttt{us.engagement\_count} in the \texttt{ORDER BY} clause, the actual cause is the preceding JOIN clause that joined a table without this column. To identify such root causes, we trace the reference relationships between clauses.

To trace those relationships, we utilize the defined logical order of SQL execution. In SQL, clauses executed in the order, with later clauses referencing tables and columns defined by earlier clauses. We trace these references backward along the execution order to find the root $C_{err}$. 
We first parse the error message to extract the table or column that caused the error, and identify the clause containing it. We then trace backward through the preceding clauses, adding any clause that references or defines the error-causing element. All identified clauses form $C_{err}$.
Based on the identified $C_{err}$, we assign rewards as follows:
\begin{equation}
r_c = \begin{cases}
-1.5 & \text{if } c \in \mathcal{C}_{\text{err}} \\
-0.5 & \text{otherwise}
\end{cases}
\end{equation}
Since the query failed to execute, we assign weak penalties even to clauses outside $C_{err}$ to discourage generating invalid SQL overall.
A detailed case study is provided in Appendix~\ref{C.3}.



\subsection{Execution-based Clause-level Policy Optimization} \label{3.3}
We assign the clause-level rewards computed in Section~\ref{3.2}, into clause-level policy optimization. Since SQL execution provides deterministic feedback on correctness, we directly assign clause-level rewards without relative comparison across multiple samples. 
Prior methods assign the query-level reward to all tokens in a generated SQL query. In contrast, our method assigns rewards at the clause level: all tokens within the same clause receive the same reward $r_c$, while tokens in different clauses receive different rewards based on their clause's correctness.
We define the policy optimization loss as:
\begin{equation}
\mathcal{L} = -\mathbb{E}\left[\sum_{c \in O} \sum_{t_i \in c} \left( r_c - \beta D_{KL}^{(i)} \right) \log \pi_\theta(t_i | t_{<i})\right]
\end{equation}
where $O$ is the generated SQL query as a sequence of clauses, $c$ denotes each clause in $O$, $t_i$ denotes a token belonging to clause $c$, $\pi_\theta(t_i|t_{<i})$ is the probability of generating token $t_i$ given previous tokens, and $D_{KL}^{(i)}$ is the KL divergence penalty to prevent the policy from deviating too far from the reference model.

Through the clause-level optimization, tokens in erroneous clause ($c \in C_{err}$) receive negative rewards, decreasing their generation probability, while tokens in correct clauses receive positive rewards, reinforcing their generation. This enables the model to selectively correct erroneous clauses while preserving correct ones.


\begin{table*}[t]
\centering
\begin{adjustbox}{max width=\linewidth}
\setlength{\tabcolsep}{12pt}
\begin{tabular}{llccc}
\toprule
\multicolumn{1}{c}{\textbf{Methods}} & \multicolumn{1}{c}{\textbf{Base Model}} & \textbf{Spider (Dev)}& \textbf{Spider (Test)} & \textbf{BIRD (Dev)} \\

\midrule
\rowcolor{gray!15}
\multicolumn{5}{l}{\textit{SFT-based methods}} \\
OmniSQL-7B~\citep{omnisql}&  Qwen2.5-Coder-7B & 85.5 & 88.9 & 66.1 \\
ROUTE~\citep{route} &  Qwen2.5-Coder-7B & 84.7 & 85.1 & 66.7 \\
SENSE-7B~\citep{SENSE} & CodeLLaMA-7B & 83.2 & 83.5 & 51.8 \\
DTS-SQL~\citep{dtssql} & DeepSeek-7B & 85.5 & 84.4 & 55.8 \\
\midrule
\rowcolor{gray!15}
\multicolumn{5}{l}{\textit{Prompt-based methods}} \\
DAIL-SQL~\citep{dailsql} & GPT-4 & 83.1 & 86.6 & 54.8 \\
MCS-SQL~\citep{mcssql} & GPT-4 & 89.5 & 89.6 & 63.4 \\
Alpha-SQL~\citep{alpha-sql}  & Qwen2.5-Coder-7B & 84.0 & - & 66.8 \\
CHASE-SQL~\citep{chasesql}  & Gemini-1.5-Pro & - & 87.6 & 73.0 \\
CHESS-SQL~\citep{chess} &  Deepseek-33B & - & 87.2 & 65.0 \\
SGU-SQL~\citep{sgusql} &  GPT-4 & 87.9 & - & 61.8 \\
\midrule
\rowcolor{gray!15}
\multicolumn{5}{l}{\textit{RL-based methods}} \\
SQL-R1~\citep{sqlr1} &  Qwen2.5-Coder-7B & 87.6 & 88.7 & 66.6 \\
SQL-R1~\citep{sqlr1} &  Qwen2.5-Coder-14B & 86.7 & 88.1 & 67.1 \\
Reward-SQL~\citep{rewardsql} &  Qwen2.5-Coder-7B & 81.7 & - & 68.9 \\
Reasoning-SQL~\citep{reasoningsql} & Qwen2.5-Coder-7B & 78.7 & - & 64.0 \\
Reasoning-SQL~\citep{reasoningsql} & Qwen2.5-Coder-14B & 81.4 & - & 65.3 \\
Graph-Reward-SQL~\citep{graphrewardsql} & Qwen2.5-Coder-7B & 81.6 & - & 63.0 \\
Arctic-SQL-R1~\citep{arctic} & Qwen2.5-Coder-7B & 87.3$^*$ & 88.8 & 68.9 \\
Arctic-SQL-R1~\citep{arctic} & Qwen2.5-Coder-14B & - & 89.4 & 70.1 \\
ExCoT~\citep{excot} & Qwen2.5-Coder-32B & - & 85.1 & 68.2 \\
\midrule
\textbf{EXPO-SQL (ours)}  & Qwen2.5-Coder-7B & \textbf{88.5} & \textbf{89.1} & \textbf{71.3} \\
\textbf{EXPO-SQL (ours)}  & Qwen2.5-Coder-14B & \textbf{89.2} & \textbf{89.5} & \textbf{73.0} \\
\bottomrule
\end{tabular}
\end{adjustbox}
\caption{Performance comparison with baselines on Spider and Bird benchmarks. We use execution accuracy (\%) as our primary metric, which measures whether the predicted SQL produces the same execution result as the gold SQL. All baseline results are reported from the original papers ($\ast$ : reproduced by us).
}
\label{tab:main_result}
\end{table*}

\section{Experiment Setups}

\subsection{Datasets}
We adopt the SynSQL-Complex-5k dataset for training following SQL-R1~\citep{sqlr1}, which consists of 5k complex NL-SQL pairs sampled from the widely-used SynSQL-2.5M~\citep{omnisql}.

For evaluation, we use Spider~\citep{spider} and BIRD~\citep{bird}, two widely-used cross-domain Text-to-SQL benchmarks. We report results on Spider development set, Spider test set, and BIRD development set. 
We also further evaluate on Spider-DK~\citep{spider-dk}, Spider-Syn~\citep{spider-syn}, and Spider-Realistic~\citep{spider-realistic}.
Dataset statistics and details are provided in Appendix~\ref{B.1}


\subsection{Baselines}
We comprehensively compare EXPO-SQL with SFT-based, prompting-based, and state-of-the-art RL-based methods. Detailed descriptions of each baseline are provided in Appendix~\ref{B.2}. 


\subsection{Implementation Details}
We use Qwen2.5-Coder 7B and 14B~\citep{qwen2.5-coder} as our base model. 
Following prior work~\citep{chess, sqlr1, omnisql}, we represent database schema using CREATE TABLE statements. 
For fair comparison, we evaluate using zero-shot inference on datasets independent of training data.
Training and inference details are provided in Appendix~\ref{B.3}.
\section{Experiment Results}

\subsection{Overall Results}
Table~\ref{tab:main_result} presents the performance comparison of various Text-to-SQL methods. EXPO-SQL consistently achieves the best performance across all benchmarks for both 7B and 14B, highlighting the effectiveness of our clause-level reward approach.

Our method shows substantial improvements over SFT-based methods across all benchmarks. With 7B parameters, it outperforms ROUTE, the best-performing SFT method, by 4.6\%p on BIRD-Dev. 
For prompting-based methods that employ multi-agent pipelines or iterative refinement, EXPO-SQL achieves comparable or superior results with significantly fewer parameters.
%
Among RL-based methods, our method outperforms all baselines with both 7B and 14B models by providing clause-level rewards that enable fine-grained supervision. Specifically, the 7B model outperforms Arctic-SQL-R1 7B by 2.4\%p and EXPO-SQL 14B outperforms Arctic-SQL-R1 14B by 2.9\%p on BIRD-Dev. The improvement is more pronounced on BIRD, which contains more complex queries requiring multiple clauses. Notably, our 7B and 14B models outperform the 32B ExCoT by 3.1\%p and 4.8\%p respectively, achieving superior performance with substantially fewer parameters.

\begin{table}[t]
\centering
\begin{adjustbox}{max width=\linewidth}
\begin{tabular}{lccccc}
\toprule
\textbf{Method} & \textbf{Detail} &\textbf{Spider-DK} & \textbf{Spider-Syn} & \textbf{Spider-Realistic} \\ \midrule
Qwen2.5-Coder-7B & - & 67.9 & 70.2 & 75.4 \\ \midrule
OmniSQL-7B & SFT & 77.8 & 69.6 & 78.0 \\
SENSE-7B & SFT & 77.9 & 72.6 & 82.7 \\ \midrule
ACT-SQL & Prompt & 68.2 & 67.9 & 75.8 \\
MAC-SQL & Prompt & 71.4 & 72.5 & 79.9 \\ \midrule
SQL-R1  & RL & 78.1 & 76.7 & 83.3 \\
Reasoning-SQL & RL &  73.3 & 69.3 & - \\ 
\textbf{EXPO-SQL (Ours)} & RL & \textbf{79.9}  & \textbf{83.1} & \textbf{83.4}  \\ \bottomrule
\end{tabular}
\end{adjustbox}
\caption{Execution accuracy (\%) on three Spider-series datasets. The existing prompting-based methods, ACT-SQL~\citep{actsql} and MAC-SQL~\citep{macsql}, use GPT-4 as the base model, while all other baselines use Qwen2.5-Coder-7B.}
\label{tab:spider_series}
\end{table}

Table~\ref{tab:spider_series} presents execution accuracy on Spider-series datasets, which evaluate generalization under domain shift (Spider-DK), synonym transformations (Spider-Syn), and realistic user queries (Spider-Realistic). EXPO-SQL outperforms existing methods across all three datasets, demonstrating the generalizability of our approach.

\subsection{Ablation Study}

\begin{table}[t]
\centering
\begin{adjustbox}{max width=0.7\linewidth}
\begin{tabular}{lc}
\toprule
\textbf{Configuration} & \textbf{BIRD (Dev)} \\ \midrule
EXPO-SQL & \textbf{71.3} \\
\quad w/o Execution Error  & 69.6 (\textcolor{red}{-1.7}) \\
\quad w/o Incorrect Result & 68.9 (\textcolor{red}{-2.4}) \\
\quad w/o Both  & 68.5 (\textcolor{red}{-2.8}) \\
\bottomrule
\end{tabular}

\end{adjustbox}
\caption{Ablation study on clause-level reward design on BIRD dev. ``w/o'' denotes replacing clause-level rewards with uniform query-level rewards for that case.}
\label{tab:ablation_clause}
\end{table}

We conduct an ablation study on the clause-level reward design on BIRD-Dev. Since correct results assign uniform positive rewards to all clauses, we ablate the clause-level rewards for incorrect result and execution error cases by replacing them with uniform query-level rewards.
As shown in Table~\ref{tab:ablation_clause}, applying clause-level rewards for both cases show the best performance. Removing clause-level rewards for execution error results in 1.7\%p drop, while removing them for incorrect result leads to a larger drop of 2.4\%p. The incorrect result case benefits more from clause-level rewards as it requires identifying specific erroneous clauses within a partially correct query. Removing both results in the largest drop of 2.8\%p confirming that clause-level rewards provide more fine-grained learning signals than query-level rewards.



\section{Further Analysis} \label{6}
For in-depth analysis of our method, we conduct additional experiments. 
We provide case-wise component analysis (Section~\ref{6.1}), comparison across various model scales (Section~\ref{6.2}), analysis by SQL complexity (Section~\ref{6.3}), effectiveness across backbone models (Section~\ref{6.4}), training efficiency analysis (Section~\ref{6.5}), reward sensitivity analysis (Section~\ref{6.6}), and statistical significance (Section~\ref{6.7}).

\begin{table}[t]
\centering
\begin{adjustbox}{max width=0.9\linewidth}
\begin{tabular}{lc}
\toprule
\textbf{$C_{err}$ Identification Method} & \textbf{BIRD (Dev)} \\
\midrule
\rowcolor{gray!15}
\multicolumn{2}{l}{\textit{Query-level reward}} \\
None & 68.9 \\ \midrule
\rowcolor{gray!15}
\multicolumn{2}{l}{\textit{Clause-level reward}} \\
+ Diff Types & 69.8 (\textcolor{blue}{+0.9}) \\
+ Clause-wise Incr. Exec. & 70.2 (\textcolor{blue}{+1.3}) \\
+ Diff Types, Clause-wise Incr. Exec. & \textbf{71.3} (\textcolor{blue}{+2.4}) \\
\bottomrule
\end{tabular}
\end{adjustbox}
\caption{Component analysis on incorrect result case. We compare methods for identifying $C_{err}$: using diff types alone, using clause-wise incremental execution (Incr. Exec.) alone, and combining both.}
\label{tab:ablation_case2}
\end{table}

\begin{table}[t]
\centering
\begin{adjustbox}{max width=0.95\linewidth}
\begin{tabular}{lc}
\toprule
\textbf{$C_{err}$ Identification Method} & \textbf{BIRD (Dev)} \\
\midrule
\rowcolor{gray!15}
\multicolumn{2}{l}{\textit{Query-level reward}} \\
None & 69.6 \\ \midrule
\rowcolor{gray!15}
\multicolumn{2}{l}{\textit{Clause-level reward}} \\
+ Error Message Parsing  & 70.1 (\textcolor{blue}{+0.5})\\
+ Error Message Parsing, Clause-level Tracing & \textbf{71.3} (\textcolor{blue}{+1.7}) \\
\bottomrule
\end{tabular}
\end{adjustbox}
\caption{Component analysis on execution error case. We compare methods for identifying $C_{err}$: using error message parsing alone, and adding clause-level tracing to find root causes.}
\label{tab:ablation_case3}
\end{table}

\subsection{Case-wise Component Analysis} \label{6.1}
We analyze the effectiveness of each component in the clause-level reward design for each case described in Section~\ref{3.2}.

\paragraph{Incorrect Result.}
We apply diff\_type detection and clause-wise incremental execution to identify $C_{err}$. Each can be used independently: diff\_type can identify $C_{err}$ directly when the matching is unique to one clause, while incremental execution can trace each clause's actual change without relying on diff type matchings. As shown in Table~\ref{tab:ablation_case2}, using diff types alone achieves 0.9\%p higher over query-level reward, while clause-wise incremental execution alone achieves 1.3\%p improvement. Combining both yields the best performance of 71.3\%, 2.4\%p improvement as diff types narrow down candidates while incremental execution pinpoints the exact clause.

\paragraph{Execution Error.}
We identify $C_{err}$ through error message parsing and clause-level tracing. Error message parsing indicates where execution failed, while clause-level tracing finds root causes in preceding clauses. As shown in Table~\ref{tab:ablation_case3}, error message parsing alone achieves 0.5\%p improvement over query-level reward. Adding clause-level tracing yields 71.3\%, 1.7\%p improvement, as it traces back to the actual source of errors.


These results confirm that each component of our clause-level reward design contributes to more accurate $C_{err}$ identification, leading to more effective clause-level rewards.

\subsection{Comparison across Various Model Scales} \label{6.2}

\begin{figure}[t]
  \includegraphics[width=\columnwidth]{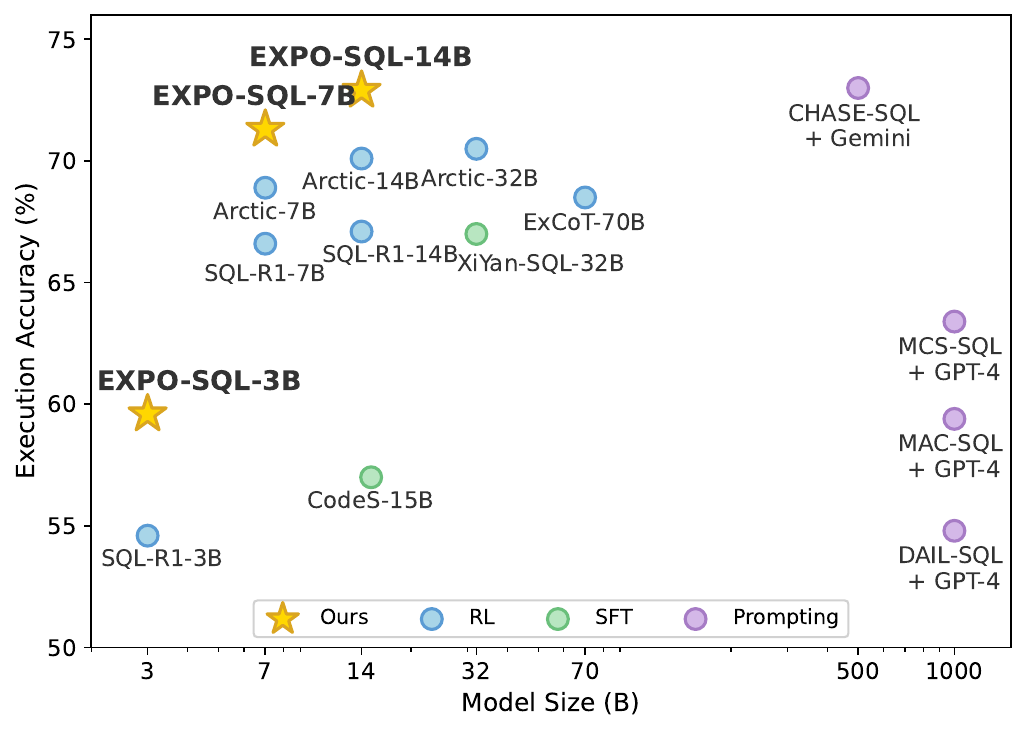}
  \caption{Execution accuracy across model scales on BIRD-Dev. EXPO-SQL achieves superior performance with significantly smaller models compared to existing RL, SFT, and prompting-based methods.}
  \label{fig:experiments}
\end{figure}

Figure~\ref{fig:experiments} analyzes the relationship between model size and performance. EXPO-SQL consistently outperforms both RL-based and SFT-based methods at identical model sizes. Furthermore, EXPO-SQL-7B achieves higher accuracy than prompting-based methods using much larger LLMs (GPT-4). These results demonstrate that clause-level rewards provide more effective learning signals, enabling smaller models to achieve comparable or superior performance with substantially fewer parameters.


\subsection{Analysis by SQL complexity} \label{6.3}

\begin{table}[t]
\centering
\begin{adjustbox}{max width=0.8\linewidth}
\begin{tabular}{lccc}
\toprule
\textbf{Method} & \textbf{Simple} & \textbf{Moderate} & \textbf{Challenging}  \\
\midrule
PPO & 74.3 & 62.8 & 58.3  \\
GRPO &74.2 & 63.1 & 58.3 \\
\midrule
\textbf{EXPO-SQL} & \textbf{76.2}& \textbf{63.9} & \textbf{63.9}  \\
\bottomrule
\end{tabular}
\end{adjustbox}
\caption{Execution accuracy (\%) of different difficulty levels on BIRD-Dev. All methods use the same reward values (ours); EXPO-SQL assigns rewards at clause-level while PPO and GRPO assign at query-level.}
\label{tab:ablation_difficulty}
\end{table}
Table ~\ref{tab:ablation_difficulty} compares EXPO-SQL with PPO and GRPO on BIRD-Dev subsets of varying difficulty. EXPO-SQL outperforms the baselines across all levels, with notable 5.6\%p  improvement on challenging queries. This demonstrates that clause-level rewards are particularly effective for complex SQL, as they enable the model to identify and correct specific erroneous clauses rather than penalizing the entire query. Further details on how complex SQL structures (nested subqueries, CTEs, set operations) are handled are provided in Appendix~\ref{A.4}.

\subsection{Effectiveness Across Various Backbone Models} \label{6.4}

\begin{table}[t]
\centering
\begin{adjustbox}{max width=\linewidth}
\begin{tabular}{lccc}
\toprule
\textbf{Model} & \textbf{Spider (Dev)} & \textbf{Spider (Test)} & \textbf{BIRD (Dev)}
\\ \midrule
Qwen2.5-Coder-7B & 83.5 & 81.5 & 51.5 \\
\hspace*{1em} w/ EXPO-SQL & \textbf{88.5} (\textcolor{blue}{+5.0}) & \textbf{89.1} (\textcolor{blue}{+7.6}) & \textbf{71.3} (\textcolor{blue}{+19.8}) \\
\midrule
Qwen2.5-Coder-14B & 83.8 & 84.8 & 56.9 \\
\hspace*{1em} w/ EXPO-SQL & \textbf{89.2} (\textcolor{blue}{+5.4}) & \textbf{89.5} (\textcolor{blue}{+4.7}) & \textbf{72.9} (\textcolor{blue}{+16.0}) \\

\midrule
DeepseekCoder-6.7B   & 78.2 & 78.5 & 58.5 \\
\hspace*{1em} w/ EXPO-SQL & \textbf{80.4} (\textcolor{blue}{+2.2}) & \textbf{81.2} (\textcolor{blue}{+2.7}) & \textbf{60.7} (\textcolor{blue}{+2.2}) \\
\midrule
Ministral-8B  & 61.7 & 62.5 & 45.9 \\
\hspace*{1em} w/ EXPO-SQL & \textbf{69.4} (\textcolor{blue}{+7.7})  & \textbf{69.7} (\textcolor{blue}{+7.2}) & \textbf{48.4} (\textcolor{blue}{+2.5}) \\
\midrule
llama3.1-8B   & 72.4 & 73.1 & 49.7 \\
\hspace*{1em} w/ EXPO-SQL & \textbf{76.7} (\textcolor{blue}{+4.3}) & \textbf{77.7} (\textcolor{blue}{+4.6})& \textbf{58.6} (\textcolor{blue}{+8.9}) \\
\bottomrule
\end{tabular}
\end{adjustbox}
\caption{Generalization across different base models. We compare execution accuracy (\%) of various base models with EXPO-SQL on three evaluation datasets.}
\label{tab:various_base}
\end{table}

\begin{table}[t]
\centering
\begin{adjustbox}{max width=\linewidth}
\begin{tabular}{lccc}
\toprule
\textbf{Model} & \textbf{Spider (Dev)} & \textbf{Spider (Test)} & \textbf{BIRD (Dev)}
\\ \midrule
OmniSQL-7B  & 85.5 & \textbf{88.9} & 66.1 \\
\hspace*{1em} w/ SQL-R1   & \textbf{87.6} (\textcolor{blue}{+2.1}) & 88.7 (\textcolor{red}{-0.2}) & 66.6 (\textcolor{blue}{+0.5}) \\
\hspace*{1em} w/ EXPO-SQL   & 87.2 (\textcolor{blue}{+1.7}) & 88.2 (\textcolor{red}{-0.7}) & \textbf{72.5} (\textcolor{blue}{+6.4}) \\
\midrule
OmniSQL-14B & 86.2 & 88.3 & 65.9 \\
\hspace*{1em} w/ SQL-R1 & 86.4 (\textcolor{blue}{+0.2}) & 87.6 (\textcolor{red}{-0.7}) & 66.6 (\textcolor{blue}{+0.7}) \\
\hspace*{1em} w/ EXPO-SQL & \textbf{87.9} (\textcolor{blue}{+1.7}) & \textbf{88.7} (\textcolor{blue}{+0.4}) & \textbf{73.2} (\textcolor{blue}{+7.3}) \\
\bottomrule
\end{tabular}
\end{adjustbox}
\caption{Performance comparison of RL methods applied to OmniSQL  (SFT model). OmniSQL is fine-tuned from Qwen2.5-Coder on synsql-2.5M.} 
\label{tab:sft_rl}
\end{table}
We verify whether EXPO-SQL achieves consistent improvements across various backbone models. Table~\ref{tab:various_base} shows improvement when EXPO-SQL is applied to different pre-trained models~\citep{deepseekcoder,mistral,llama}. Our method achieves consistent improvements across all benchmarks regardless of model type and size.

Table~\ref{tab:sft_rl} shows the results when applying RL methods to OmniSQL, which is fine-tuned from Qwen2.5-Coder on SynSQL-2.5M. On BIRD-Dev, EXPO-SQL achieves 6.4\%p (7B) and 7.3\%p (14B) improvement, significantly outperforming SQL-R1. This demonstrates that clause-level rewards are effective regardless of the base model.

\subsection{Training Efficiency} \label{6.5}

\begin{table}[t]
\centering
\begin{adjustbox}{max width=0.9\linewidth}
\begin{tabular}{lccc}
\toprule
\textbf{Metric} & \textbf{GRPO} & \textbf{PPO} & \textbf{EXPO-SQL} \\
\midrule
Generation time (s) & 15.59 & 11.46 & 10.16 \\
Reward time (s) & 0.69 & 0.72 & 0.60 \\
Actor update (s) & 6.61 & 6.61 & 6.66 \\
Other (ref / critic) (s) & 1.15 & 2.65 & 1.16 \\
\midrule
\textbf{Total step time (s)} & \textbf{25.67} & \textbf{21.45} & \textbf{20.02} \\
\midrule
DB calls / sample & 2.0 & 2.0 & $\sim$3.4 \\
DB time / batch (s) & $\sim$0.38 & $\sim$0.38 & $\sim$0.52 \\
DB \% of step time & 1.5\% & 1.8\% & 2.6\% \\
\bottomrule
\end{tabular}
\end{adjustbox}
\caption{Wall-clock time breakdown per training step (8$\times$ H100, Qwen2.5-Coder-7B, batch size 64).}
\label{tab:training_efficiency}
\end{table}

Table~\ref{tab:training_efficiency} compares the wall-clock time per training step. Despite performing more database calls per sample for clause-wise incremental execution, EXPO-SQL is 22\% faster than GRPO and 7\% faster than PPO. The speedup comes from sample efficiency: GRPO requires $n{=}8$ samples per prompt for group-relative optimization, while EXPO-SQL uses REINFORCE++~\citep{hu2025reinforce++} with $n{=}1$ sample per prompt, saving 5.43s in generation time. Compared to PPO, EXPO-SQL requires no critic model, eliminating $\sim$1.5s per step for value network updates. The extra DB calls from clause-wise execution add less than 3\% to total step time, as each call takes only $\sim$3ms.

\begin{figure}[t]
    \centering
    \includegraphics[width=0.9\columnwidth]{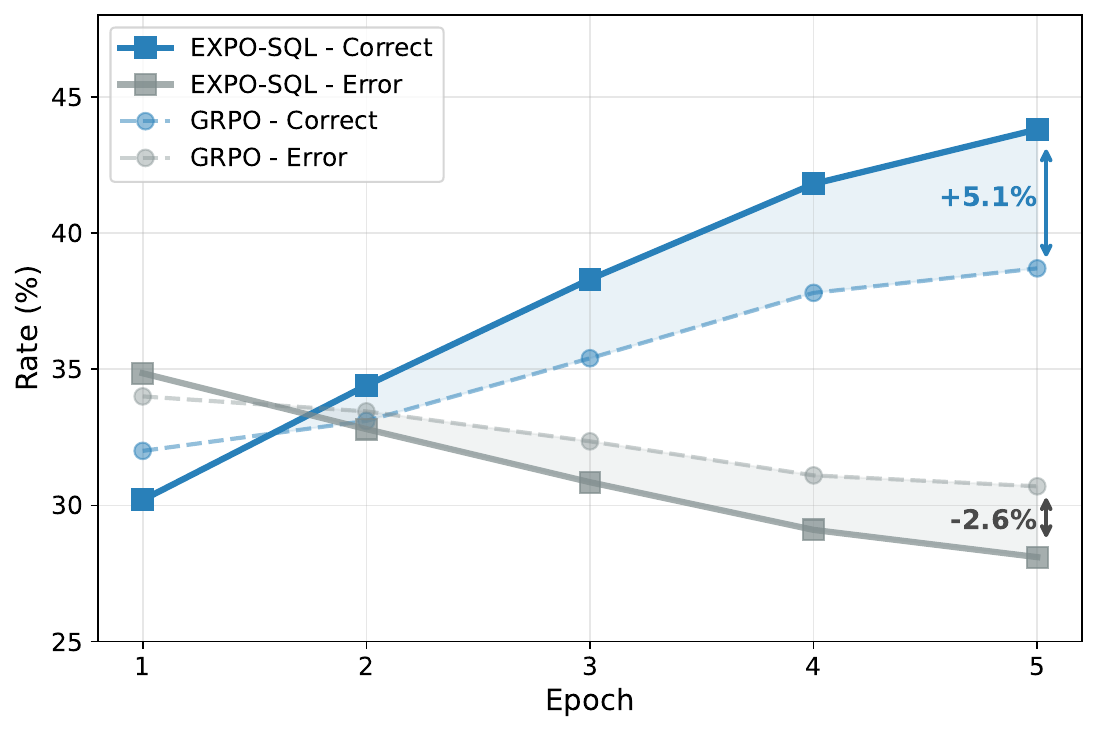}
    \caption{Correct and error rates during training for EXPO-SQL and GRPO. Correct rate represents the ratio of queries producing correct execution results. Error rate denotes the average of incorrect result and execution error rates.}
    \label{fig:training_dynamics}
\end{figure}

Beyond wall-clock efficiency, Figure~\ref{fig:training_dynamics} shows that EXPO-SQL also achieves better \emph{learning} efficiency. Throughout training, EXPO-SQL consistently maintains a higher correct rate and lower error rate than GRPO, achieving a 5.1\%p higher correct rate and 2.6\%p lower error rate at the final epoch. This indicates that clause-level rewards provide more informative gradient signals per training step, enabling the model to improve more rapidly than under query-level rewards.

\subsection{Reward Sensitivity Analysis} \label{6.6}

\begin{table}[t]
\centering
\begin{adjustbox}{max width=0.85\linewidth}
\begin{tabular}{lcc}
\toprule
\textbf{Configuration} & \textbf{Gap Scale} & \textbf{BIRD (Dev)} \\
\midrule
No differentiation (uniform) & $\times$0 & 68.5 \\
\midrule
Halved & $\times$0.5 & 70.8 \\
Ours (default) & $\times$1.0 & \textbf{71.3} \\
Doubled & $\times$2.0 & 70.9 \\
\bottomrule
\end{tabular}
\end{adjustbox}
\caption{Reward sensitivity analysis. We scale the clause-level reward gap while preserving the hierarchical structure. Performance is robust across scales (70.8--71.3), while removing differentiation entirely drops to 68.5.}
\label{tab:reward_sensitivity}
\end{table}

Table~\ref{tab:reward_sensitivity} examines the sensitivity of performance to reward magnitude. We scale the gap between $C_{err}$ and non-$C_{err}$ clause rewards while preserving the hierarchical structure (Match $>$ Mismatch $>$ Execution Error). All differentiated configurations fall within a 0.5-point range (70.8--71.3), while removing clause-level differentiation entirely drops performance to 68.5. This confirms that the structure of separating erroneous from correct clauses drives performance, not the specific magnitudes.

\subsection{Statistical Significance} \label{6.7}
We conducted paired bootstrap resampling (10,000 iterations) comparing EXPO-SQL against the uniform query-level reward baseline. The improvements are statistically significant: BIRD-Dev ($p < 0.001$) and Spider-Dev ($p < 0.05$). Additionally, consistent gains across five independent backbone architectures (Table~\ref{tab:various_base}), where all 15 combinations of backbone $\times$ benchmark show positive improvement, further support that the effect is not due to random variance.





\section{Conclusion}
We proposed EXPO-SQL, a novel reinforcement learning framework for Text-to-SQL.
This method enabled clause-level policy optimization by leveraging execution results to identify erroneous clauses and assign clause-level rewards. 
To the best of our knowledge, this is the first work to introduce clause-level rewards in RL-based Text-to-SQL.
Extensive experiments showed that clause-level supervision enables more precise and fine-grained learning.


\section*{Limitations}
We demonstrate that leveraging execution feedback enables effective clause-level reward assignment for Text-to-SQL. However, our experiments are conducted on SQLite-based benchmarks, and generalization to other database dialects (e.g., PostgreSQL, MySQL) requires further investigation. Additionally, error message formats vary across database systems, which require adaptation of our parsing approach for different executors.


\section*{Acknowledgments}
This work was partly supported by Institute of Information \& communications Technology Planning \& Evaluation(IITP) grant funded by the Korea government(MSIT) (RS-2019-II190421, 13\%; IITP-2026-RS-2024-00437633, 12\%; RS-2025-25442569, 12\%; IITP-2026-RS-2024-00360227, 13\%; and No.RS-2023-00228970, 12\%) and National Research Foundation of Korea(NRF) grant funded by the Korea government(MEST) (RS-2024-00352717, 13\%), funded by the Korea government(MSIT) (RS-2025-00521391, 13\%), and funded by the Ministry of Education (RS-2025-25433088, 12\%)
\bibliography{custom}

\clearpage
\appendix
\section*{Appendix}

\begin{table*}[t]
\centering

\small
\begin{tabular}{l|l|p{6cm}|p{4cm}}
\toprule
\textbf{Level} & \textbf{Diff Type} & \textbf{Natural Language Description} & \textbf{Potential Erroneous Clauses} \\
\midrule
\multirow{2}{*}{\textbf{Column}} 
 & \texttt{col\_count} & The number of columns in the output schema is incorrect. & \texttt{SELECT}, \texttt{FROM}, \texttt{JOIN} \\
 & \texttt{col\_name} & Column count is correct, but the wrong attributes were projected. & \texttt{SELECT}, \texttt{FROM}, \texttt{JOIN} \\
\midrule
\multirow{7}{*}{\textbf{Row}} 
 & \texttt{row\_order} & Rows are correct, but the sorting sequence is misaligned. & \texttt{ORDER BY} \\
 & \texttt{row\_dedup} & Unique rows match, but the use of \texttt{DISTINCT} or grouping is incorrect. & \texttt{SELECT (DISTINCT)}, \texttt{GROUP BY} \\
 & \texttt{row\_subset} & The result is missing rows due to overly restrictive conditions. & \texttt{WHERE}, \texttt{JOIN}, \texttt{HAVING}, \texttt{LIMIT} \\
 & \texttt{row\_superset} & Extra rows are generated (e.g., unintended Cartesian products). & \texttt{JOIN}, \texttt{WHERE} (OR/IN) \\
 & \texttt{row\_emptied} & The query returns no data, while the gold result is non-empty. & \texttt{WHERE}, \texttt{JOIN} (No match), \texttt{HAVING} \\
 & \texttt{row\_created} & Rows are returned for a query that should yield an empty result. & \texttt{FROM}, \texttt{JOIN} \\
 & \texttt{row\_disjoint} & Predicted and gold rows have zero intersection (completely wrong data). & \texttt{FROM}, \texttt{JOIN}, \texttt{WHERE} \\
\midrule
\textbf{Value} 
 & \texttt{row\_partial} & Structure aligns, but calculated values (aggregates/math) are wrong. & \texttt{SELECT} (Aggregates, Expressions) \\
\bottomrule
\end{tabular}
\caption{Comprehensive taxonomy of SQL result discrepancies. This classification assumes an alias-free environment and maps each diff type to its structural or logical origins in SQL clauses. Our incremental execution strategy resolves the 1:N ambiguity between diff types and clauses by capturing the exact moment a discrepancy first manifests.}
\label{tab:diff_types}
\end{table*}

\section{Additional Details About EXPO-SQL} \label{A}

\subsection{Difference Type Classification} \label{A.1}

We classify the differences between predicted execution result $R_{pred}$ and gold result $R_{gold}$ into 10 diff\_type. Since SQL results are returned as tables, any difference appear as changes in columns, rows, or values.

\paragraph{Theoretical Intractability of Unique Identification.}
Uniquely identifying a single erroneous clause from execution results alone is theoretically intractable, as it reduces to the SQL equivalence problem, which is undecidable in the general case~\citep{chu2017hottsql} and remains open even for conjunctive queries under the bag semantics that SQL uses~\citep{khamis2020bag, marcinkowski2024bag}.
Execution-based evaluation on finite database instances is also inherently incomplete for determining semantic equivalence~\citep{zhong2020semantic}.
Instead of targeting unique identification, we narrow down the set of candidate erroneous clauses by leveraging the operational semantics of individual SQL clauses, providing substantially finer-grained learning signals than query-level rewards that penalize the entire query uniformly.

\paragraph{Derivation from SQL Semantics.}
A SQL result is a bag of tuples over a fixed schema, with optional ordering~\citep{guagliardo2017formal, green2007provenance}. Any difference between two SQL results must therefore manifest in one of four axes: \emph{schema}, \emph{content}, \emph{ordering}, or \emph{value}. Table~\ref{tab:diff_axes} shows how these axes yield our 10 diff types.

\begin{table}[h]
\centering
\small
\begin{tabular}{l|l}
\toprule
\textbf{Axis} & \textbf{Diff Types} \\
\midrule
Schema & \texttt{col\_count}, \texttt{col\_name} \\
Content (bag) & \texttt{row\_subset}, \texttt{row\_superset}, \\
              & \texttt{row\_disjoint}, \texttt{row\_emptied}, \\
              & \texttt{row\_created}, \texttt{row\_dedup} \\
Ordering & \texttt{row\_order} \\
Value & \texttt{row\_partial} \\
\bottomrule
\end{tabular}
\caption{Mapping of SQL result axes to diff types. Schema differences follow from the relational model, bag-content differences enumerate all pointwise multiplicity relations~\citep{green2007provenance}, ordering is separate when \texttt{ORDER BY} is present~\citep{zhong2020semantic}, and value-level differences cover cell-wise mismatches.}
\label{tab:diff_axes}
\end{table}

This decomposition aligns with the dichotomy between why and why-not provenance~\citep{chapman2009whynot, huang2008provenance} and with prior SQL testing and evaluation work~\citep{chandra2015xdata, miao2019explaining, zhong2020semantic}.

\paragraph{Column-Level Differences.}
Column-level differences occur when the schema of the two results differs. If the number of columns differs, we classify it as \texttt{col\_count}. If the column count matches but the column names differ, we classify it as \texttt{col\_name}. These differences typically originate from the \texttt{SELECT} clause.

\paragraph{Row-Level Differences.}
Row-level differences occur when the schema matches but the rows differ. We further distinguish these based on how the row sets relate to each other.
If both results contain the same rows but in different order, we classify it as \texttt{row\_order}, which typically originates from \texttt{ORDER BY}.
If the unique rows match but duplicate counts differ, we classify it as \texttt{row\_dedup}, typically caused by \texttt{DISTINCT} or \texttt{GROUP BY}.
For differences in row existence, we consider five cases: \texttt{row\_subset} when the predicted result is missing some rows, \texttt{row\_superset} when extra rows are present, \texttt{row\_emptied} when the predicted result is empty but gold is not, \texttt{row\_created} when the predicted result has rows but gold is empty, and \texttt{row\_disjoint} when the two results have no intersection. These typically originate from \texttt{WHERE}, \texttt{JOIN}, or \texttt{HAVING} clauses.

\paragraph{Value-Level Differences.}
Value-level differences occur when the row structure aligns but specific cell values differ. We classify this as \texttt{row\_partial}, which typically indicates errors in aggregate functions or arithmetic expressions in the \texttt{SELECT} clause.

Table~\ref{tab:diff_types} provides the full mapping between diff types and potential erroneous clauses.

\paragraph{Why Each Type Matters.}
Each diff type signals the \emph{direction} of the error.
\texttt{row\_subset} (missing rows) indicates an over-restrictive clause, while \texttt{row\_superset} (extra rows) indicates an under-restrictive one.
These map to different candidate clauses: \texttt{row\_subset} $\rightarrow$ \{\texttt{WHERE}, \texttt{JOIN}, \texttt{HAVING}, \texttt{LIMIT}\}, \texttt{row\_superset} $\rightarrow$ \{\texttt{JOIN}, \texttt{WHERE}\}.
Merging them loses this direction.
This asymmetry is consistent with prior work on why vs. why-not provenance~\citep{chapman2009whynot, huang2008provenance}, SQL mutation testing~\citep{chandra2015xdata}, and Text-to-SQL evaluation~\citep{zhong2020semantic}.

\subsection{Clause-wise Incremental Execution Algorithm} \label{A.2}

Algorithm~\ref{alg:incremental_main} describes the procedure for identifying $C_{err}$ in the incorrect result case with clause-wise incremental execution. The algorithm consists of three phases: preprocessing, incremental execution, and erroneous clause identification.

\paragraph{Preprocessing.}
We first replace the \texttt{SELECT} clause with \texttt{SELECT *} to isolate the effect of each clause during incremental execution. When the query contains \texttt{GROUP BY}, \texttt{SELECT *} violates SQL constraints, so we process \texttt{SELECT} and \texttt{GROUP BY} together as a single unit. We then reorder the clauses according to the logical execution order.

\paragraph{Incremental Execution.}
We execute clauses incrementally following the logical execution order: \texttt{FROM/JOIN} $\rightarrow$ \texttt{WHERE} $\rightarrow$ \texttt{GROUP BY} $\rightarrow$ \texttt{HAVING} $\rightarrow$ \texttt{SELECT} $\rightarrow$ \texttt{ORDER BY} $\rightarrow$ \texttt{LIMIT}. For the first clause (\texttt{FROM/JOIN}), there is no previous result to compare, so we compute $d_1 = \text{diff}(R_{gold}, R_1)$ by comparing directly with $R_{gold}$. For subsequent clauses ($i \geq 2$), we compute $d_i = \text{diff}(R_{i-1}, R_i)$ by comparing the results before and after adding each clause.

\paragraph{Erroneous Clause Identification.}
After obtaining all $d_i$ through incremental execution, we compute the final difference $D^* = \text{diff}(R_{gold}, R_{pred})$. A clause $c'_i$ is identified as erroneous and added to $C_{err}$ if its corresponding $d_i$ appears in $D^*$, meaning the change introduced by $c'_i$ contributed to the final difference.

\subsection{SQLite Execution Error Taxonomy} \label{A.3}
\label{appendix:sqlite_errors}

When an execution error occurs, our system parses the error string to identify the faulty components. Table~\ref{tab:sqlite_error_list} categorizes common SQLite error messages and defines how they serve as triggers for the clause-level tracing mechanism described in Section~\ref{exec_error}.

\begin{table*}[t]
\centering

\small
\begin{tabular}{p{1.5cm}|p{6.5cm}|p{6.5cm}}
\toprule
\textbf{Category} & \textbf{Common SQLite Error Patterns} & \textbf{Detailed Description} \\
\midrule
\textbf{Schema Reference} & 
\texttt{no such column: [col\_name]} \newline
\texttt{no such table: [table\_name]} \newline
\texttt{ambiguous column name: [col\_name]} & 
Occurs when the query attempts to access an object not present in the database or the current subquery context. These typically indicate a failure in the data retrieval stage (\texttt{FROM}/\texttt{JOIN}). \\
\midrule
\textbf{Logical Misuse} & 
\texttt{misuse of aggregate function [func]()} \newline
\texttt{aggregate functions are not allowed in...} \newline
\texttt{misuse of aliased identifier} & 
Triggered when an operation violates the semantic rules of SQL execution order, such as using an aggregate function in a prohibited clause (e.g., \texttt{WHERE}). \\
\midrule
\textbf{Data \& Constraint} & 
\texttt{datatype mismatch} \newline
\texttt{NOT NULL constraint failed: [col]} \newline
\texttt{UNIQUE constraint failed: [col]} & 
Indicates that the query structure is correct, but the operation violates data integrity or involves incompatible data types during calculation. \\
\midrule
\textbf{Syntax Error} & 
\texttt{near "[token]": syntax error} \newline
\texttt{unrecognized token: "[char]"} \newline
\texttt{incomplete input} & 
Occurs when the SQL engine cannot parse the query string due to grammatical failures, preventing the formation of a logical execution plan. \\
\bottomrule

\end{tabular}
\caption{Classification of common SQLite execution errors. These messages are utilized to extract key identifiers (e.g., column/table names) as initial triggers for the clause-level tracing process.}
\label{tab:sqlite_error_list}
\end{table*}

The error messages listed in Table~\ref{tab:sqlite_error_list} act as diagnostic signals for identifying $C_{err}$. The system employs different strategies based on the error category:

\paragraph{Schema and Constraint Tracing.} 
For schema reference errors, the system extracts the identifier (e.g., \texttt{age} from \texttt{no such column: age}) and performs a clause-level tracing through the SQL clauses to identify where this object was omitted or incorrectly joined. Similarly, constraint failures prompt an analysis of the \texttt{JOIN} or \texttt{WHERE} conditions that might have introduced invalid data.

\paragraph{Logical and Syntax Errors.} 
For logical misuse, the specific function in the message serves as an anchor to flag the clause that attempted the illegal operation. For syntax errors, the \texttt{"near [token]"} hint isolates the physical location of the error clause.

\paragraph{Clause-level Tracing.}
A \texttt{"no such column"} error is interpreted as a failure in the data flow. The tracing logic investigates whether the column was lost during a join operation or if the model failed to account for a table alias, allowing us to pinpoint the root-cause clause even if the error was caught later in the execution sequence (e.g., in \texttt{ORDER BY}).

\subsection{Handling Complex SQL Structures} \label{A.4}

For complex SQL structures, we attribute errors at the granularity of the enclosing clause, as summarized in Table~\ref{tab:complex_sql}.
Even without decomposing internal structures, this still distinguishes correct and erroneous clauses: the 5.6\%p improvement on Challenging queries (Table~\ref{tab:ablation_difficulty}) confirms its effectiveness.

\begin{table}[t]
\centering
\small
\begin{tabular}{p{2cm}|p{5cm}}
\toprule
\textbf{Structure} & \textbf{Attribution Strategy} \\
\midrule
Nested subquery & Handled as part of its enclosing clause. \\
\midrule
Set operations (UNION / INTERSECT / EXCEPT) & Split at set operation boundaries; each sub-query is analyzed independently. \\
\midrule
Window function & Treated as part of \texttt{SELECT}; errors appear as \texttt{row\_partial} at the \texttt{SELECT} step. \\
\midrule
Alias & Resolved to source tables/columns before incremental execution. \\
\midrule
CTE & Identified as erroneous when the referencing \texttt{FROM/JOIN} step produces a diff; internal decomposition is skipped. \\
\bottomrule
\end{tabular}
\caption{Attribution strategies for complex SQL structures.}
\label{tab:complex_sql}
\end{table}

\section{Additional Details About Experimental Setups} \label{B}

\subsection{Datasets} \label{B.1}

\paragraph{Training.}
SynSQL-Complex-5K is a subset of SynSQL-2.5M~\citep{omnisql}, containing 5K complex NL-SQL pairs. The queries involve multiple joins, nested subqueries, and aggregations.

\paragraph{Evaluation.}
Spider~\citep{spider} contains 10,181 question-SQL pairs across 200 databases covering 138 domains (dev: 1,034, test: 2,147). BIRD~\citep{bird} contains 12,751 pairs from 95 large-scale databases across 37 specialized domains (dev: 1,534). For robustness evaluation, we use Spider-DK~\citep{spider-dk} which requires domain knowledge, Spider-Syn~\citep{spider-syn} which replaces schema words with synonyms, and Spider-Realistic~\citep{spider-realistic} which removes explicit schema mentions.

\subsection{Baselines} \label{B.2}
SFT-based methods fine-tune models on text-SQL pairs, including SENSE~\citep{SENSE}, DTS-SQL~\citep{dtssql} which decomposes the task into subtasks, and OmniSQL~\citep{omnisql} which synthesizes chain-of-thought reasoning data. 
Prompting-based methods leverage LLMs without training, including DAIL-SQL~\citep{dailsql} with in-context learning, MCS-SQL~\citep{mcssql} with execution-based selection, and CHASE-SQL~\citep{chasesql} with multi-agent refinement. 
RL-based methods optimize with execution feedback, including SQL-R1~\citep{sqlr1}, ReasoningSQL~\citep{reasoningsql} with partial rewards, and Arctic-SQL-R1~\citep{arctic}.

\subsection{Implementation Details} \label{B.3}
We implement EXPO-SQL using the Ray + vLLM framework~\citep{hu2024openrlhf} for efficient inference, and extend the RL++ framework\citep{hu2025reinforce++} to support clause-level reward propagation. We train for 5 epochs with batch size 64, learning rate 1e-6, and temperature 0.8 on 8 NVIDIA H100 80GB GPUs. The KL penalty coefficient $\beta$ is set to 0.01. We use AdamW optimizer with cosine annealing scheduler and warmup ratio 0.1. For inference, we follow SQL-R1~\citep{sqlr1} and use self-consistency with 8 samples. All experiments are conducted with Python 3.10 and CUDA 12.0+.


\section{Additional Experimental Results} \label{C}

\subsection{Per-Clause Error Rate Dynamics} \label{C.clause_dynamics}

Since each incorrect-result case produces a set of causal clauses $C_{err}$, we can track the \emph{causal error rate} per clause type, the fraction of mismatch cases in which a given clause appears in $C_{err}$, as a diagnostic view of how error distribution shifts during training on the SynSQL-Complex-5K training set. Table~\ref{tab:clause_dynamics} compares the rates between the early (Epoch 1--2) and late (Epoch 5) training stages.

\begin{table}[t]
\centering
\begin{adjustbox}{max width=\linewidth}
\begin{tabular}{lccc}
\toprule
\textbf{Clause} & \textbf{Early (Ep 1--2)} & \textbf{Late (Ep 5)} & \textbf{Change} \\
\midrule
FROM & 90.1\% & 82.1\% & $-$8.0\,pp \\
SELECT & 72.7\% & 64.7\% & $-$8.0\,pp \\
JOIN & 58.9\% & 67.4\% & $+$8.5\,pp \\
GROUP BY & 47.9\% & 39.7\% & $-$8.2\,pp \\
WHERE & 32.9\% & 26.2\% & $-$6.7\,pp \\
ORDER BY & 15.9\% & 14.7\% & $-$1.2\,pp \\
LIMIT & 8.3\% & 6.0\% & $-$2.3\,pp \\
\bottomrule
\end{tabular}
\end{adjustbox}
\caption{Causal error rate per clause type during training on SynSQL-Complex-5K. Values represent the fraction of mismatch cases in which each clause appears in $C_{err}$.}
\label{tab:clause_dynamics}
\end{table}

Most clauses show consistent reduction in their causal rate, with FROM, SELECT, and GROUP BY each dropping by around 8\,pp. JOIN increases ($+$8.5\,pp) not because JOIN performance worsens, but because as the model resolves earlier syntax errors (the unexecutable rate drops from 26.8\% to 7.5\%), previously failing queries become executable and expose underlying JOIN errors that were previously masked. Such diagnostic signals, unavailable under query-level rewards, reveal that easy errors are resolved first while harder cases persist, providing insight into how execution-based RL training progresses.

\subsection{Case Study: Clause-wise Incremental Execution Example} \label{C.2}
\begin{figure*}[!t]
  \centering
  \includegraphics[width=\linewidth]{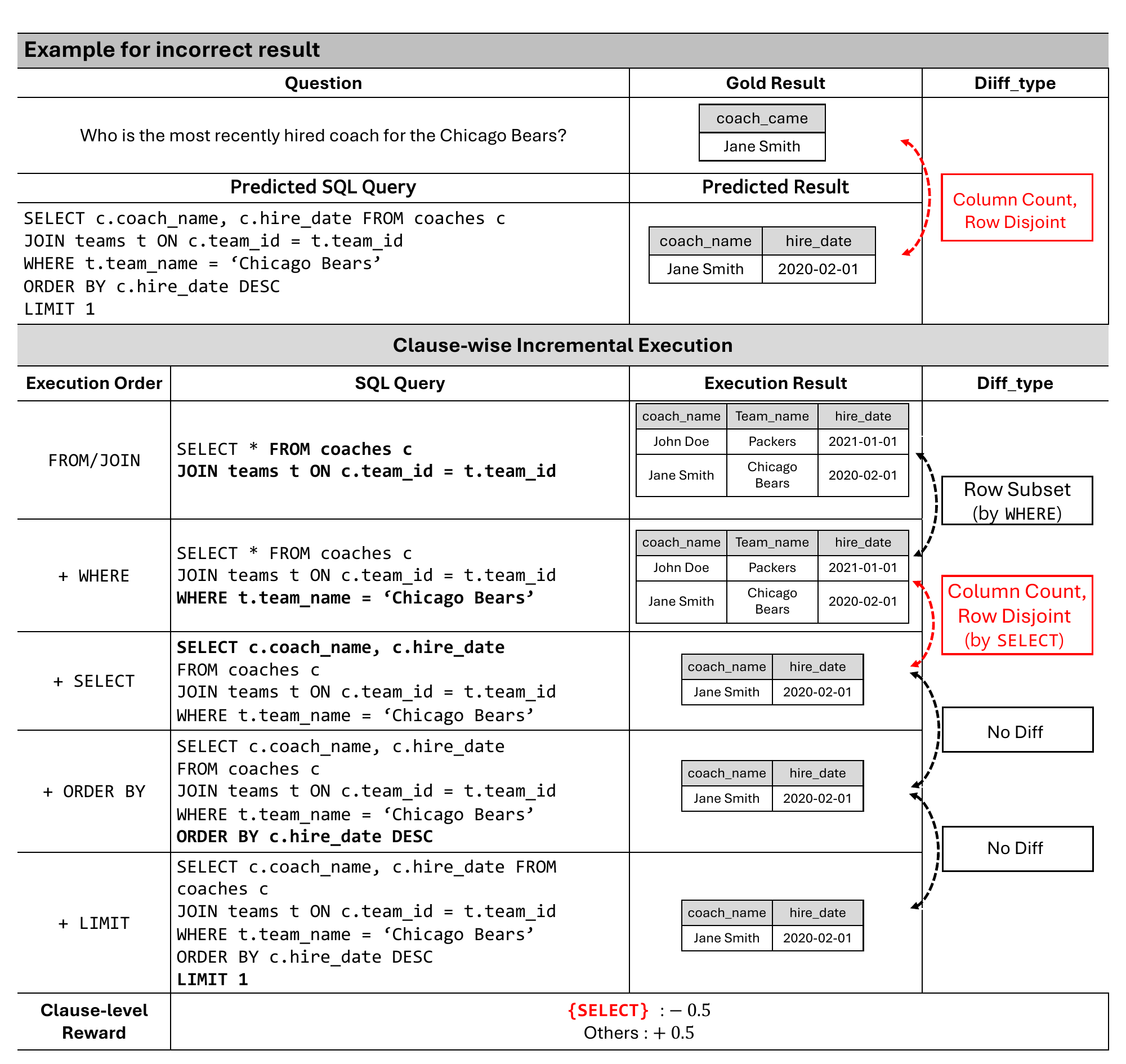}
  \caption {Case study of clause-wise incremental execution - (1)}
    \label{result_err_case_1}
\end{figure*}

\begin{figure*}[!t]
  \centering
  \includegraphics[width=\linewidth]{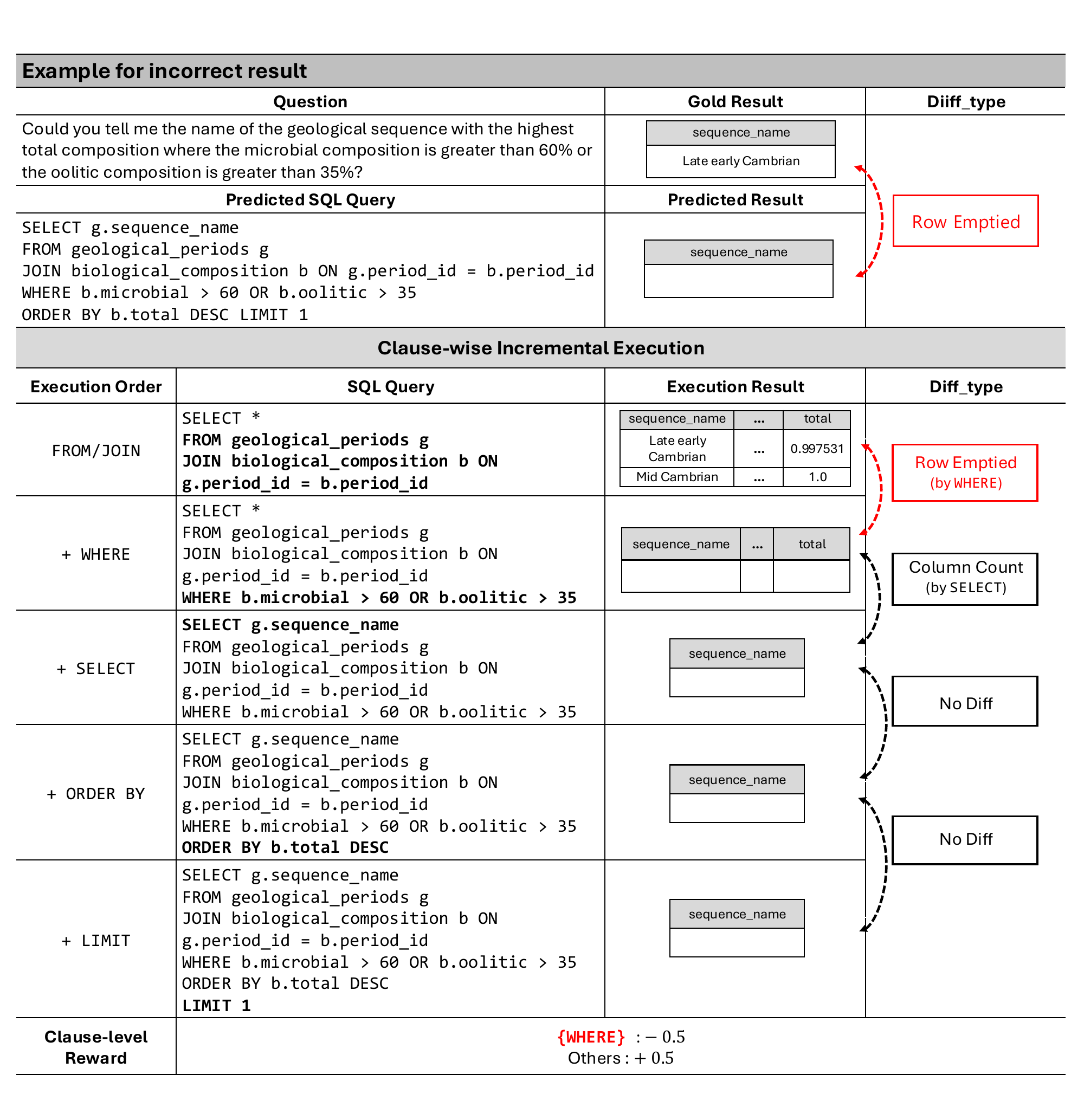}
  \caption {Case study of clause-wise incremental execution- (2)}
    \label{result_err_case_2}
\end{figure*}

Figure~\ref{result_err_case_1} and Figure~\ref{result_err_case_2} illustrate how clause-wise incremental execution identifies erroneous clauses when the query produces an incorrect result. By incrementally executing each clause and comparing intermediate results, we can trace which clause introduced the difference observed in the final result.

\subsection{Case Study: Clause-level Tracing for Execution Errors} \label{C.3}
\begin{figure*}[!t]
    \centering
    \includegraphics[width=0.8\linewidth]{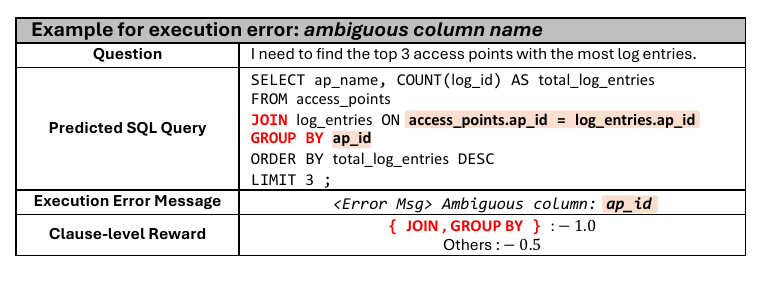}
    \caption {Case study of clause-level tracing: ambiguous column error}
    \label{exe_err_case_1}
\end{figure*}

\begin{figure*}[!t]
    \centering
    \includegraphics[width=0.8\linewidth]{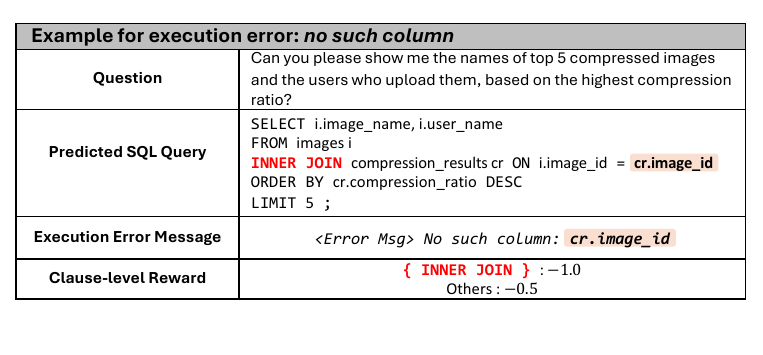}
    \caption {Case study of clause-level tracing: column not found error}
    \label{exe_err_case_2}
\end{figure*}

Figure~\ref{exe_err_case_1} and Figure~\ref{exe_err_case_2} demonstrate how clause-level tracing identifies erroneous clauses from database error messages. Figure~\ref{exe_err_case_1} shows an ambiguous column error case, and Figure~\ref{exe_err_case_2} shows a column not found error case.

\clearpage

\end{document}